\documentclass[letterpaper,twocolumn,10pt]{article}
\usepackage{usenix-2020-09}

\usepackage{tikz}
\usepackage{amsmath}

\usepackage{filecontents}
\usepackage[skins]{tcolorbox} 
\newtcolorbox{hint}{
    colback=black!2!white,
    colframe=black!80!white,
    colbacktitle=black!50!white,
    fonttitle=\bfseries,
    title=Hints,
    enhanced,
    attach boxed title to top center={yshift=-2mm},
}
\newtcolorbox{discuss}{
    colback=brown!2!white,
    colframe=brown!95!white,
    colbacktitle=brown!75!white,
    fonttitle=\bfseries,
    title=Discuss this,
    enhanced,
    attach boxed title to top center={yshift=-2mm},
}
\makeatletter

\newcommand{\Comment}[2]{\textit{\textbf{\textcolor{cyan!65!black}{#1}}: \textcolor{gray!65!black}{#2}}}

\newcommand{\OUTLINE}[1]{{\Comment{outline}{#1}}}

\newcommand{\GS}[1]{{\Comment{GS}{[#1]}}}

\colorlet{revision}{blue!70!black}

\newcommand\red[1]{\textcolor{red}{#1}}
\newcommand\blue[1]{\textcolor{blue}{#1}}

\newcommand{\allnotes}[1]{}
\renewcommand{\allnotes}[1]{\textit{#1}}

\definecolor{commentcolor}{RGB}{107,142,35}

\newcommand{\codecomment}[1]{\textrm{\textit{\textcolor{commentcolor}{#1}}}}

\newcommand{\sys}{TARDIS\xspace}

\newcommand{\cf}{constant folding\xspace}

\usepackage{amsfonts}
\usepackage{enumitem,kantlipsum}
\usepackage{graphicx}
\usepackage{fancybox}

\usepackage[hang,flushmargin]{footmisc}
\usepackage{booktabs}   
\usepackage{subcaption} 
\usepackage[normalem]{ulem}
\usepackage[export]{adjustbox}                                    

\usepackage{outlines}

\usepackage{listings}
\usepackage{breakurl}                                             
\usepackage{multicol}

\usepackage{setspace}
\usepackage[linesnumbered,boxed,lined,ruled,vlined, nofillcomment]{algorithm2e}

\usepackage{filecontents}
\usepackage{cleveref}

\usepackage{multirow}
\usepackage{array}
\usepackage{balance}
\usepackage{rotating}

\usepackage{titlesec}
\usepackage{pgfplots}
\usepackage{bchart}
\usepackage{pdfpages}
\graphicspath{{gnuplot/img/}}

\usepackage{listings}
\usepackage{longtable}

\definecolor{codegreen}{rgb}{0,0.6,0}
\definecolor{codegray}{rgb}{0.5,0.5,0.5}
\definecolor{codepurple}{rgb}{0.58,0,0.82} 
\definecolor{backcolour}{rgb}{0.95,0.95,0.92}

\begin{document}

\date{}

\title{Accelerating Large Language Models through Partially Linear Feed-Forward Network}

\author{
Gansen Hu\textsuperscript{1} \and 
Zhaoguo Wang\textsuperscript{2} \and 
Jinglin Wei\textsuperscript{3} \and
Wei Huang\textsuperscript{4} \and
Haibo Chen\textsuperscript{5}\\
\em Institute of Parallel and Distributed Systems, SEIEE, Shanghai Jiao Tong University\\
\{\textsuperscript{1}hugansen,
\textsuperscript{2}zhaoguowang,
\textsuperscript{3}weijinglin,
\textsuperscript{5}haibochen\}@sjtu.edu.cn,
\textsuperscript{4}boogiepop021230@gmail.com
}

\maketitle

\begin{abstract}
    Large language models (LLMs) demonstrate remarkable capabilities but face deployment challenges due to their massive parameter counts. While existing compression techniques like pruning can reduce model size, it leads to significant accuracy degradation under high compression ratios. We present a novel perspective inspired by constant folding in compiler optimization. Our approach enables parameter reduction by treating activation functions in LLMs as linear functions.

    However, recent LLMs use complex non-linear activations like GELU that prevent direct application of this technique. We propose \sys, which enables optimization of LLMs with non-linear activations by \textit{partially} approximating them with linear functions in frequently occurring input ranges. For outlier inputs, \sys employs an online predictor to dynamically fall back to original computations.

    Our experiments demonstrate that \sys achieves 80\% parameter reduction in feed-forward networks, while significantly outperforming state-of-the-art pruning methods Wanda and RIA with up to 65\% higher accuracy. In practical deployments for a 7B model, \sys achieves 1.6× end-to-end inference speedup when integrated with the vLLM serving system, and 1.4× speedup with the widely adopted HuggingFace implementation, while incurring only a 10.9\% accuracy trade-off.

\end{abstract}
    
\section{Introduction}
\label{sec:intro}


Recently, transformer-based large language models (LLMs) have revolutionized natural language processing, demonstrating remarkable capabilities across diverse tasks. Deploying these models for inference is notoriously challenging due to their enormous number of parameters~\cite{wang2024model,xu2024survey,zhou2024survey}, often reaching billions or even trillions~\cite{googlesw98:online}. To address this, various techniques have been proposed to compress LLMs, including pruning~\cite{sun2023simple,ma2023llm,zhang2024plug}, quantization~\cite{frantar2022optq,lin2024awq,yao2022zeroquant}, knowledge distillation~\cite{gu2023knowledge,timiryasov2023baby}, and low-rank approximation~\cite{li2023losparse,saha2023matrix,chand2023dsformer}.
While pruning has gained popularity due to its simplicity~\cite{yangllmcbench}, it often leads to significant accuracy degradation under high compression ratios - a 80\% compression could result in up to 70\% accuracy loss.

Inspired by the success of \cf~\cite{Constant13:online} in traditional compiler optimizations, we propose a novel approach to LLM compression that maintains high accuracy even under aggressive compression ratios.
Constant folding, a fundamental compiler optimization technique that reduces both program code size and runtime computation costs, has been successfully employed across numerous programming languages including C/C++~\cite{Optimize87:online}, Python~\cite{pythoncp39:online}, and SQL~\cite{SparkSQL40:online}.
We discover a unique opportunity to adapt this technique for LLM compression.

The architecture of an LLM primarily consists of two core components: the Multi-Head Attention (MHA) block and the feed-forward network (FFN) block, with the FFN block accounting for 67\% to 80\%~\cite{almazrouei2023falconseriesopenlanguage} of the total parameters.
The FFN block comprises two matrix multiplications separated by a non-linear activation function (e.g., ReLU or GELU).

Our key observation is that through linear approximation of the non-linear activation function, we can leverage the associative property of matrix multiplication to merge the two FFN matrices into a single, compact matrix.
This transformation enables a reduction in FFN parameters by up to 87.5\% in theory (detailed in \Cref{sec:opportunity}).

However, modern LLMs predominantly use sophisticated non-linear activation functions, particularly 
GELU~\cite{hendrycks2016gaussian} (adopted in GPT series models~\cite{brown2020language}) 
and SiLU~\cite{elfwing2018sigmoid} (used in LLaMA2~\cite{touvron2023llamaopenefficientfoundation}).
These complex functions are crucial for capturing intricate patterns in the data, and naively approximating them with linear functions leads to severe performance degradation.
Our experiments with a 7B model show that a simple linear approximation of GELU results in a 75\% drop in model accuracy.

In this paper, we present \sys, a novel approach that enables constant folding in LLMs with non-linear activation functions at low approximation error.
Our key insight is that the inputs to these activation functions in LLMs are typically concentrated within a narrow range.
This distribution pattern allows us to \textit{partially} approximate the non-linear activation function with a linear function in the "hot" input ranges (where most values occur), introducing minimal approximation error.
For outlier inputs that fall outside these ranges, \sys employs an efficient online predictor to identify them and dynamically falls back to the original activation function computation.
Extensive experiments demonstrate that \sys reduces the FFN weight matrices by up to 80\%, while significantly outperforming state-of-the-art pruning methods in accuracy. Compared to leading pruning methods like Wanda and RIA, \sys achieves up to 65\% higher accuracy on downstream tasks. For inference, \sys delivers up to 1.6× end-to-end speedup on the state-of-the-art vLLM serving system~\cite{10.1145/3600006.3613165}, and 1.4× speedup on HuggingFace's widely-used implementation~\cite{Transf37:online}.

The key contributions of this paper are:
\begin{itemize}
    \item We identify a novel opportunity to apply constant folding to LLMs. Our method treats activation functions as partially linear functions, reducing FFN block parameters by up to 87.5\% in theory.
    \item We introduce \sys, a new system for constant folding in modern LLMs. It handles non-linear activation functions through partial linear approximation and includes efficient fallback mechanisms.
    \item We conduct extensive experiments with \sys. Results show up to 80\% actual parameter reduction. This significantly outperforms current state-of-the-art pruning approaches.
\end{itemize}

\section{Background and Motivation}
In this section, we first cover the basics of LLM inference and highlight the performance bottlenecks introduced by massive parameters. We then review existing approaches to reducing LLM size and identify their constraints.
\begin{figure}
    \centering
    \begin{subfigure}{0.47\columnwidth}
        \includegraphics[width=\textwidth]{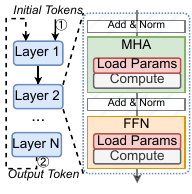}
        \caption{The inference process of 
        a transformer based LLM}
        \label{fig:transformer-inference-process}
    \end{subfigure}
    \hfill
    \centering
    \begin{subfigure}{0.50\columnwidth}
        \includegraphics[width=\textwidth]{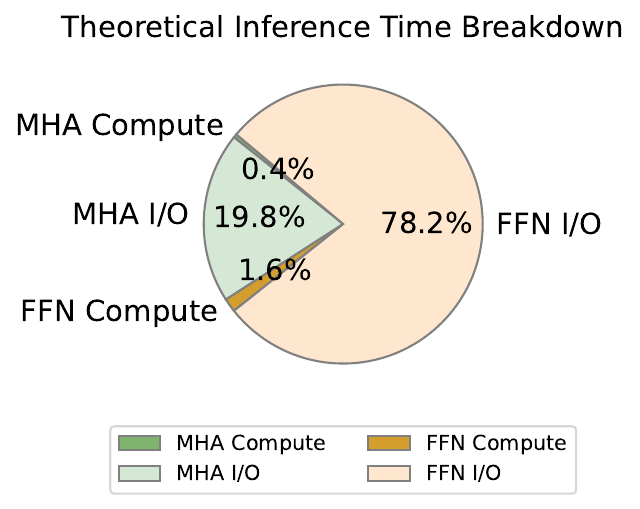}
        \caption{Theoretical Breakdown of the inference time on an RTX 4090}
        \label{fig:llm-inference-breakdown}
    \end{subfigure}
    \caption{The inference process of transformer based LLM (a), where initial tokens are user prompts, output tokens are response from LLM;
    the inference time breakdown (b) with 91 initial tokens, 178 output tokens for Falcon-7B (Other operations takes <0.5\% time and not shown).}
    \label{fig:transformer-inference}
\end{figure}
\subsection{Large Language Model Inference}
\label{sec:llm_inference}
\textbf{LLM Inference Overview.} LLM inference generates output tokens in response to input tokens through two main phases.
In the first phase, as shown in \Cref{fig:transformer-inference-process}, the model processes all initial input tokens through its $N$ layers sequentially to establish context. 
In the next phase, the model generates output tokens one at a time in an auto-regressive manner, where each generated token becomes part of the input sequence for predicting the next token. This process continues until either the desired output length is reached or a stop condition is met.

\noindent
\textbf{Operations in LLM Inference.}
The transformer architecture, which forms the backbone of LLMs, consists of two main components: the multi-head attention (MHA) block and the feed-forward network (FFN) block.
The MHA block enables the model to focus on different parts of the input sequence simultaneously through multiple attention heads, while the FFN block, composed of two linear transformations with an activation function in between, processes the features from the attention block. As \Cref{fig:transformer-inference-process} shows, an LLM typically stacks multiple transformer layers, where each layer sequentially processes the input through its MHA and FFN blocks.
During inference, the model parameters for both blocks need to be loaded from memory before performing computations on input tokens.


\subsection{Cost of LLM Inference}
Despite their remarkable capabilities, deploying LLMs for inference tasks remains challenging due to their massive parameter counts~\cite{lin2024awq,liu2023deja,xu2024survey,wang2024model}. 
This large parameter count significantly impacts inference performance.

To quantify this impact, we analyzed the inference time breakdown of Falcon-7B using the SharedGPT dataset~\cite{philschm22:online}, which contains 90K real-world chat histories between humans and LLMs. 80\% of the total parameters are in the FFN blocks of Falcon-7B.
Using the dataset's average lengths (91 tokens for human input, 178 tokens for LLM response), 
we measured the inference time with the vLLM inference engine~\cite{vllmproj38:online} on an RTX 4090, where attention blocks take 0.9s, and the FFN blocks takes 2.1s, showing the bottleneck is in FFN blocks.
We also performed a therotical analysis of the time spent on computation and I/O operations (performed on the 1TB/s bandwidth VRAM of RTX 4090) in both MHA and FFN blocks.
\Cref{fig:llm-inference-breakdown} shows that parameter loading I/O dominates the inference time, with the FFN I/O alone consuming 78.2\% of the total time. 
This high I/O overhead stems from LLMs' auto-regressive nature: generating each token requires loading all model parameters into the hardware's cache (\Cref{fig:transformer-inference-process}), resulting in repeated, expensive I/O operations~\cite{wang2024model,xu2024survey,zhou2024survey}.
\begin{figure}[t]
    \centering
    \includegraphics[width=\columnwidth]{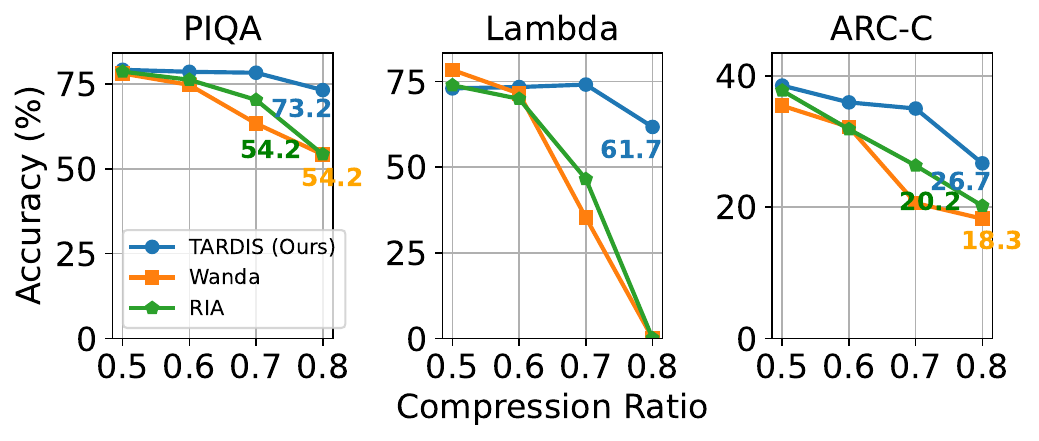}
    \caption{Accuracy of different pruning methods under different compression ratio of the FFN block. Existing baselines suffer pronounced accuracy loss on high compression ratio.}
    \label{fig:pruning_high_compression_ratio}
\end{figure}

\subsection{LLM Compression}
To reduce the I/O overhead during LLM inference, researchers have developed various compression techniques. These techniques can be broadly categorized into four approaches: pruning~\cite{sun2023simple,ma2023llm,zhang2024plug,liu2023deja,song2024powerinfer}, quantization~\cite{frantar2022optq,lin2024awq,wei2023outlier,kim2023finequant}, knowledge distillation~\cite{gu2023knowledge,timiryasov2023baby}, and low-rank approximation~\cite{li2023losparse,saha2023matrix,chand2023dsformer}. 
These approaches adopt different ideas and usually excel in different settings~\cite{yangllmcbench,kuzmin2023pruning}.
Among these approaches, pruning has emerged as one of the most popular techniques due to its conceptual simplicity~\cite{yangllmcbench}. Pruning works by identifying and removing less important parameters based on certain criteria, such as magnitude or gradient information. Users can specify a target compression ratio, and the selected parameters are set to zero, effectively reducing both storage requirements and computational costs.

While simple, pruning suffers from significant accuracy loss under high compression ratios. We evaluated two state-of-the-art pruning methods, Wanda~\cite{sun2023simple} and RIA~\cite{liu2023deja}, on Falcon-7B using three benchmarks: PIQA~\cite{bisk2020piqa}, Lambada~\cite{Eleuther75:online}, and ARC-Challenge~\cite{allenaia10:online}. \Cref{fig:pruning_high_compression_ratio} shows that at a 50\% compression ratio, the accuracy drop remains modest at 0.9\%-4.8\% across all tasks. However, when increasing the compression to 80\%, the accuracy can drop to even zero on the Lambada dataset, making the model practically unusable.


\begin{figure}[t]
    \centering
    \includegraphics[width=\columnwidth]{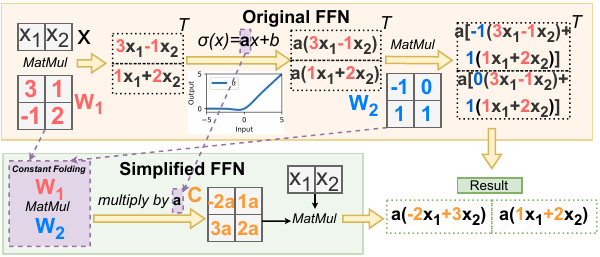}
    \caption{Illustration of constant folding in the FFN block of an LLM. The FFN contains two weight matrices \red{$W_1$}, \blue{$W_2$}. By treating activation function $\sigma$ as a linear function $ax+b$, matrix multiplications can be reordered from $a(xW_1)W_2$ to $x(aW_1W_2)$. Since $aW_1W_2$ consists of only constants, it can be pre-computed into a single matrix \textbf{\textcolor{orange}{$C$}} before inference.}
    \label{fig:mlp_constant_folding_example}
\end{figure}

\section{Opportunity and Challenge}
Inspired by the success of constant folding in compiler optimizations, we discover a novel opportunity to reduce I/O costs by simplifying FFN blocks in LLMs through activation function modification. Our approach achieves higher accuracy under aggressive compression compared to state-of-the-art pruning methods. Rather than directly manipulating model parameters like existing pruning techniques, we focus on simplifying the activation functions in FFN blocks, which enables parameter reduction through constant folding.

\begin{figure}[t]
    \centering
    \includegraphics[width=\columnwidth]{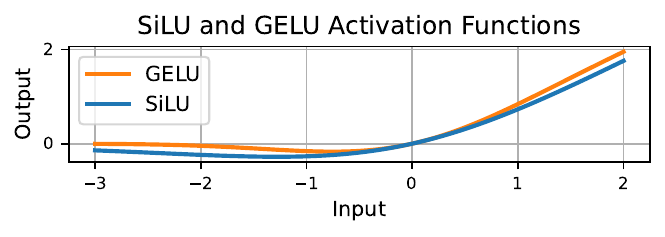}
    \caption{Figure of SiLU and GELU activation functions in a input range of \texttt{[-3, 2]}.}
    \label{fig:relu-gelu-activation-functions}
\end{figure}
\subsection{Opportunity}
\label{sec:opportunity}
A key insight for simplifying FFN blocks in LLMs is that their activation functions can be approximated as linear functions.
The FFN block in transformer models consists of two weight matrices $W_1$, $W_2$, and an activation function between them. As shown in \Cref{fig:mlp_constant_folding_example}, a simple FFN block with activation computes:
$$
\begin{aligned}
FFN(x)=\sigma(xW_1)W_2
\end{aligned}
$$
where matrix $W_1$ is of size $d\times h$ and $W_2$ is $h\times d$, $\sigma$ is the activation function. 
A function $f(x)$ is linear if and only if it can be expressed as $f(x)=ax+b$, where $a$ and $b$ are constants. 
When the activation function is approximated by a linear function, the FFN becomes the following linear transformation:
$$
FFN(x)=a((xW_1)+b\footnote{\label{note_b_presentation}To ease presentation, we omit $b$ by regarding it as 0, without lossing generalizability.})W_2
$$


This linear representation of activation function enables us to leverage matrix computation properties to optimize the FFN block. Specifically, we can reorder the computation from $a(xW_1)W_2$ to $x(aW_1W_2)$ using matrix associativity.
This reordering would not be possible with non-linear activation functions.
The key advantage is that $aW_1W_2$ contains only constants and can be pre-computed into a single matrix $C$ before inference. As illustrated in \Cref{fig:mlp_constant_folding_example}, with input \texttt{[$x_1$,$x_2$]}, the FFN computes \texttt{t=[$a(-2x_1+3x_2),a(x_1+2x_2)$]}. By pre-computing $aW_1W_2$ into matrix $C=\big(\begin{smallmatrix}-2a & a\\3a & 2a\end{smallmatrix}\big)$, we can simplify the entire FFN into a single matrix multiplication $xC$ that produces the same result \texttt{t}.
This transformation significantly reduces the parameter count in FFN blocks. 
Originally, an FFN block with weight matrices $W_1 \in \mathbb{R}^{d\times h}$ and $W_2 \in \mathbb{R}^{h\times d}$ contains $2\times d\times h$ parameters. 
After folding into a single matrix $C \in \mathbb{R}^{d\times d}$, it contains only $d^2$ parameters.
For typical LLM architectures like GPT2~\cite{openaico17:online}, BLOOM~\cite{le2023bloom}, and Falcon~\cite{almazrouei2023falconseriesopenlanguage} where $h=4d$, this reduces the parameter count by up to 87.5\% per FFN block.

\subsection{Challenge}
While constant folding in LLMs shows promise, a significant challenge arises from the widespread adoption of sophisticated non-linear activation functions in modern language models~\cite{hendrycks2016gaussian,elfwing2018sigmoid,clevert2015fast,klambauer2017self,dauphin2017language}. 
Our analysis of the top 15 text generation models on HuggingFace~\cite{Transf37:online}, a leading model repository, reveals that all models use complex activations - 7 employ GELU and 8 use SiLU.
These sophisticated activation functions make naive linear approximation challenging.

Consider GELU as an example. As illustrated in \Cref{fig:relu-gelu-activation-functions}, GELU exhibits a smooth but intricate input-output relationship.
When we attempt to approximate GELU with a simple linear function (as described in \Cref{sec:opportunity}), the model suffers substantial accuracy degradation.
To validate this, we modified Falcon-7B by replacing its GELU activation with a linear function in the FFN blocks, derived through linear regression on the original activation's input-output data.
When evaluated on the Lambada~\cite{Eleuther75:online} benchmark, the model's accuracy plummeted from 75\% to 0, making it impractical for real-world applications.

\begin{figure}[t]
    \centering
    \includegraphics[width=\columnwidth]{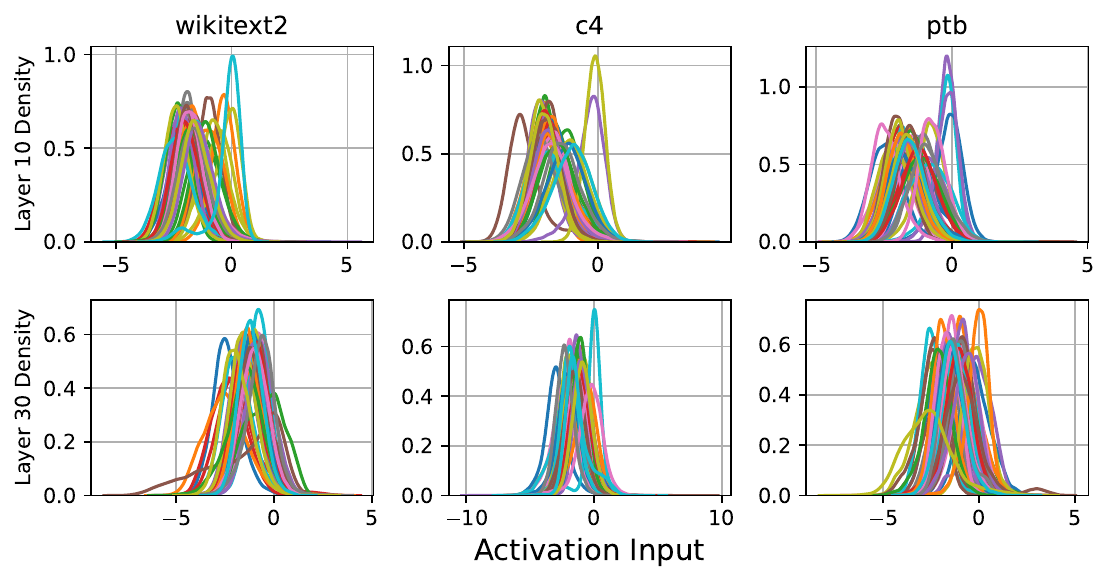}
    \caption{Density estimation of the activation function input distribution for 50 neurons of layer 10 and 30 in Falcon-7B. Each curve represents the kernel density estimation of the input distribution for a single neuron. The higher the density, the more activation inputs fall into the range. 
    The left two figures are profiled on WikiText2, the middle two on C4, and the right two on PTB.}
    \label{fig:neural-act-inputs}
\end{figure}

\section{Insights and Overview}
Through careful analysis of LLM internals, we have identified two key insights that enable effective approximation of non-linear activation functions with linear ones while preserving model accuracy.

\subsection{Insights}
\label{sec:insight}
\textbf{Insight 1. Skewed distribution of activation function inputs.} The input distribution for activation functions in LLMs exhibits a highly skewed pattern, with inputs predominantly concentrated within a narrow range of the input space. 
To validate this observation, we analyzed three widely-used language task datasets: Wikitext-2~\cite{merity2016pointer}, C4~\cite{JMLR:v21:20-074}, and Penn Treebank (PTB)~\cite{10.3115/1075812.1075835}. 
A FFN block consists of multiple \textit{neurons}, where each neuron processes input through its weights and activation function. The activation input distribution at the neuron level is crucial for understanding model behavior~\cite{liu2023deja,song2024powerinfer}. 
We examined this distribution within FFN weight matrices. Each weight matrix can be decomposed into individual neurons, which represent distinct computational units. For example, in \Cref{fig:mlp_constant_folding_example}, $W_1$ comprises two neurons represented as columns: 
$\big(\begin{smallmatrix}
    3 \\
    -1 
\end{smallmatrix}\big)$ and 
$\big(\begin{smallmatrix}
    1 \\
    2 
\end{smallmatrix}\big)$.
Using 20K random tokens from each dataset, we collected comprehensive statistics.

\Cref{fig:neural-act-inputs} illustrates the input activation distribution for 50 randomly sampled neurons from two layers of Falcon-7B across these datasets.
Our analysis reveals consistent distribution patterns within the same layer across different datasets.
Notably, approximately 65\% of activation inputs for each neuron are concentrated within just 20\% of the total input range, demonstrating significant skewness.
This pattern holds true across various model architectures.
\begin{table}[t]
    \centering
    \resizebox{\columnwidth}{!}{%
    \begin{tabular}{lllll}
    \hline
    Model Name & \begin{tabular}[c]{@{}l@{}}Activation\\ Function\end{tabular} & Wikitext-2 & C4 & PTB \\ \hline
    Falcon-7B  & GELU & 20.2\% & 20.6\% & 20.7\% \\
    Falcon-40B & GELU & 19.6\% & 19.7\% & 20.1\% \\
    BLOOMZ-7B1 & GELU & 19.1\% & 19.4\% & 19.8\% \\
    LLaMA2-7B  & SiLU & 18.4\% & 18.5\% & 18.2\% \\ \hline
    \end{tabular}%
    }
    \caption{Average percentage of activation input ranges containing 65\% of inputs relative to total range length, measured across neurons in three datasets.}
    \label{tab:skew-act-inputs}
    \end{table}
\Cref{tab:skew-act-inputs} presents the average range portion containing 65\% of activation inputs relative to the total range for four LLMs: Falcon-7B/40B, Bloomz-7B1, and LLaMA2-7B. While the first three models employ GELU activation, LLaMA2-7B uses SiLU. Across all models and datasets, this portion consistently remains between 18-20\%. These findings align with previous research~\cite{mirzadehrelu,song2024powerinfer,liu2023deja}, confirming that skewed activation input distribution is an inherent characteristic of LLMs.
This insight reveals that activation function inputs follow a power law-like distribution, enabling effective approximation of non-linear activation functions within a compact "hot" input range.

\noindent
\textbf{Insight 2. Non-uniform distribution of layer and neuron importance.} Not all layers and neurons contribute equally to the model's performance. Using the same methodology as in the previous insight, we collected the Mean Square Error (MSE) of approximating the activation function with linear functions for each neuron in each layer. Specifically, for each layer and neuron, we first identify the input range containing a certain portion (e.g., 85\%) of activation inputs based on the profiled distribution. We then approximate the activation function within this range using a linear function $y=ax+b$, where parameters $a$ and $b$ are obtained through linear regression to minimize the loss within the range. This process is repeated for different portions of inputs (from 65\% to 95\%) to analyze the approximation error patterns.

\begin{figure}
    \centering
    \begin{subfigure}{\columnwidth}
        \includegraphics[width=\columnwidth]{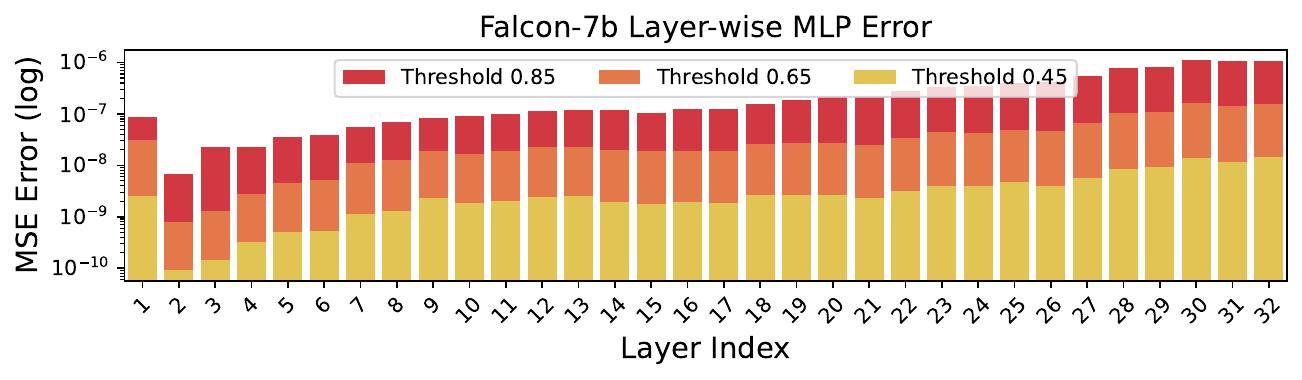}
        \caption{Layer-wise error distribution}
        \label{fig:dist-errors-layer}
    \end{subfigure}
    \hfill
    \begin{subfigure}{\columnwidth}
        \includegraphics[width=\columnwidth]{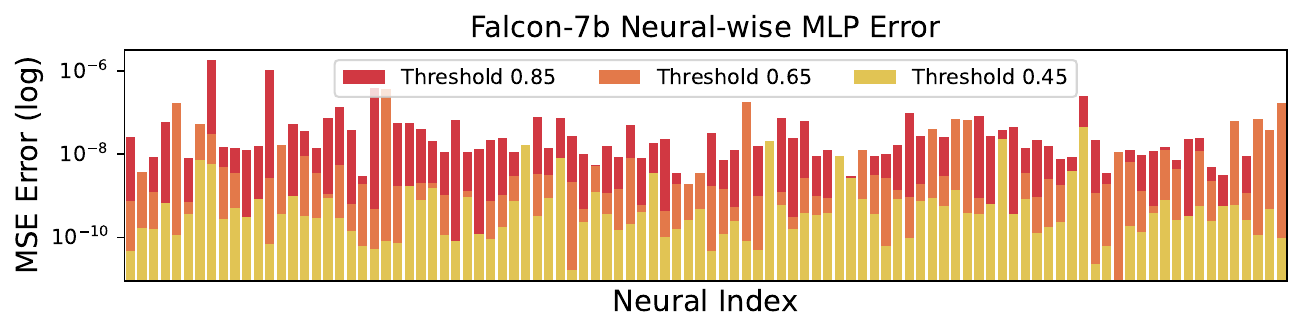}
        \caption{Neuron-wise error distribution}
    \end{subfigure}
    \caption{Error distribution of layer, and neuron in the 10-th layer in Falcon-7B. The error is profiled on the C4 dataset.}
    \label{fig:dist-errors-layer-neural}
\end{figure}

\Cref{fig:dist-errors-layer-neural}.a shows significant differences in errors across layers. For instance, when 85\% of inputs fall within the linear range, the error of the first layer ($10^{-7}$) is an order of magnitude larger than that of the second layer ($10^{-8}$). Previous works have also shown that layer importance is non-uniformly distributed~\cite{zhang2024investigating,you2024shiftaddllm,pmlr-v235-yin24e}. Similarly, neuron importance within a layer varies significantly. \Cref{fig:dist-errors-layer-neural}.b presents the error distribution for 1\% of neurons randomly selected from the first layer of Falcon-7B, showing error variations by nearly three orders of magnitude (from $10^{-10}$ to $10^{-7}$). Previous studies have also found similar patterns, where neuron activation is non-uniformly distributed, with outliers having a larger impact on model performance~\cite{song2024powerinfer,10.5555/3600270.3602468,lin2024awq}.
In summary, \Cref{fig:dist-errors-layer-neural} highlights the non-uniform distribution of layer and neuron importance, underscoring the need to treat each layer and neuron distinctly in our design approach.

\subsection{Overview}
Armed with these insights, we propose \sys, the first system that enables constant folding in LLMs while preserving model capabilities and reducing parameter size.
Our approach leverages two key observations:
1) the skewed distribution of activation inputs allows us to focus approximation on the most frequently accessed input ranges;
2) the varying importance of different layers and neurons suggests that approximation thresholds should be adapted accordingly.

\begin{figure}[t]
    \includegraphics[width=1.0\linewidth]{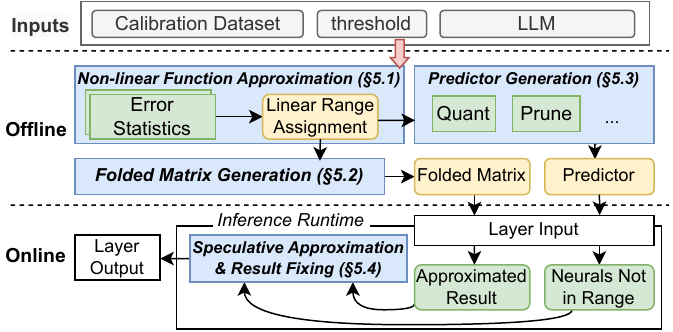}
    \caption{Architecture of \sys with offline and online components. The offline component generates the constant folded matrix and predictor based on the calibration dataset. The online component performs inference using the constant folded matrix and predictor.}
    \label{fig:overall}
\end{figure}

As illustrated in \Cref{fig:overall}, \sys takes three inputs: an LLM, a calibration dataset, and a threshold \(t\) that defines what portion of inputs should fall within the linear approximated range. 
The system generates two key outputs: a compressed model weight matrix (produced via constant folding based on approximated linear ranges) and a predictor that identifies neurons operating outside their approximated ranges.

The system architecture consists of offline and online components. The offline component performs two tasks.
First, it analyzes the calibration dataset--a small number of text samples--to collect: 1) activation input distributions for each layer and neuron, and 2) importance scores based on approximation errors when replacing activation functions with linear functions.
Based on these statistics, \sys approximates the activation function with linear functions, focusing on the ``hot'' input ranges where most activations occur.
The approximation range for each neuron is determined by its importance score, which \sys calculates based on the error introduced by linear approximation under different thresholds.
The final outputs are the constant folded matrix and a predictor for identifying out-of-range neuron activations.
During inference, the online component first performs a speculative computation using the constant folded matrix.
It then uses the predictor to identify neurons operating outside their approximated ranges.
For these identified neurons, it corrects the results by: 1) subtracting the incorrectly computed linear approximations, and 2) recomputing the actual results using the original activation function.

\begin{figure}
    \centering
    \includegraphics[width=1\linewidth]{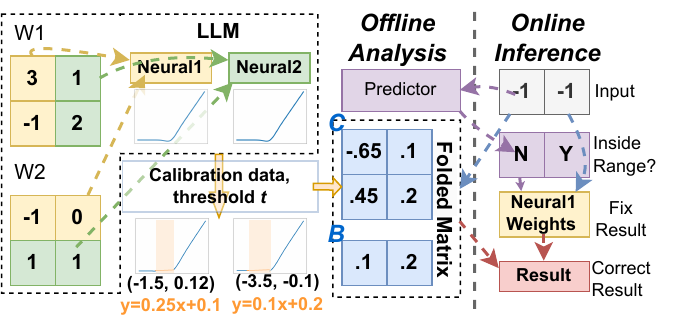}
    \caption{Example of an FFN in an LLM constant folded by \sys. The constant folded matrix is generated based on approximated linear ranges for each neuron. The predictor identifies neuron one is not in the approximated range. The online component fixes the FFN result by subtracting the wrong result and recomputing the actual result using the original activation function.}
    \label{fig:example-constant-folded-llm}
\end{figure}

\Cref{fig:example-constant-folded-llm} illustrates this process using a simple FFN with two neurons (each neuron corresponds to one column in $W_1$ and one row in $W_2$, as shown in yellow and green). The FFN uses GELU as activation function.
In the offline phase, \sys determines linear approximations for each neuron's frequent input range. For neuron one, the range is $(-1.5,0.12)$ with slope 0.25 and intercept 0.1; for neuron two, the range is $(-3.5,-0.1)$ with slope 0.1 and intercept 0.2.
These ranges are selected based on calibration data to ensure at least \(t\) proportion of inputs fall within them. The system then generates constant folded matrices \(C\) and \(B\) using these linear parameters.
When inference with input token $x=(\begin{smallmatrix}-1 & -1\end{smallmatrix}\big)$, \sys
1) performs speculative computation: $xC+B=\big(\begin{smallmatrix}0.3 & -0.1\end{smallmatrix}\big)$;
2) uses predictor to determine that only neuron two's input falls within its approximated range;
3) corrects the result by
    a) subtracting neuron one's approximate result: $(0.25x\big(\begin{smallmatrix}3 \\ -1\end{smallmatrix}\big)+0.1)\big(\begin{smallmatrix}-1 & 0\end{smallmatrix}\big)=\big(\begin{smallmatrix}0.4 & 0\end{smallmatrix}\big)$, and
    b) adding back neuron one's actual result: $GELU(x\big(\begin{smallmatrix}3 \\ -1\end{smallmatrix}\big))\big(\begin{smallmatrix}-1 & 0\end{smallmatrix}\big)$.
This correction mechanism ensures accurate results while preserving the efficiency benefits of constant folding where applicable.

\section{\sys Design}
In this section, we present the detailed design of \sys, which comprises offline and online components.
The offline component consists of three key aspects: approximating non-linear activation functions with linear functions (\Cref{sec:non_linear_approx}), generating constant folded matrices (\Cref{sec:matrix_gen}), and constructing predictors (\Cref{sec:predictor_gen}).
The online component details how these elements work together during inference (\Cref{sec:runtime}).

\subsection{Non-linear Function Approximation}
\label{sec:non_linear_approx}
\sys aims to reduce FFN parameter counts and speedup inference by replacing complex non-linear activation functions with simple linear functions where possible. 
At a high level, our non-linear activation function replacing approach works as follows.
For each FFN layer that consists of two weight matrices ($W_1$ and $W_2$) and a non-linear activation function $\sigma$ in between, we 1) break down the computation of the entire FFN into computations on individual neurons, where each neuron corresponds to a column in $W_1$ and a row in $W_2$.
2) For each neuron, analyze its input patterns to the activation function using a calibration dataset. 3) Replace the neuron's non-linear activation with linear functions ($y = ax + b$) within its most frequently observed input range. The linear functions are obtained using least squares regression.

\begin{figure}[t]
    \centering
    \includegraphics[width=0.95\columnwidth]{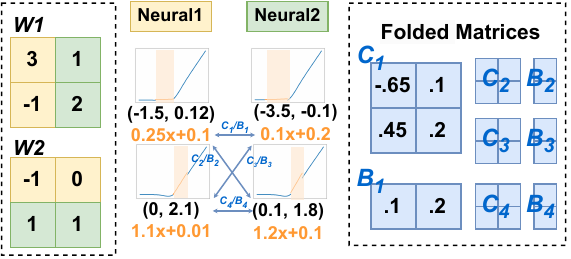}
    \caption{Example of the exponential growth in the number of folded matrices.}
    \label{fig:too_many_matrices}
\end{figure}

An intuitive design would be to replace each neuron's non-linear activation with multiple linear ranges, as different input ranges may exhibit distinct patterns. For example, for input ranges [$l_1,a$), [$a,b$), and [$b,l_2$),
we could fit three separate linear functions to better approximate the activation function. However, a significant challenge arises with this approach - when trying to fold the weight matrices, we need a separate folded matrix for each combination of ranges across neurons. With $h$ neurons and $r$ ranges per neuron, this leads to $r^h$ folded matrices. \Cref{fig:too_many_matrices} illustrates this exponential growth - even with just 2 neurons and 2 ranges each, we need 4 different folded matrices to cover all possible combinations.

Given that modern language models typically have over 10,000 neurons per FFN layer, storing multiple ranges quickly becomes impractical. This leads us to adopt a single-range strategy, which we describe in detail in the following section.

\subsubsection{Single Range Approximation Strategy}
To overcome the exponential storage overhead of multi-range approximation, \sys adopts a pragmatic single-range strategy. For each neuron, we approximate its non-linear activation using one linear function within a carefully chosen input range.

More specifically, for each neuron $n$, we aim to find a linear function $y=ax+b$ that best approximates the activation within a range $[l_1, l_2)$. The approximation function $\phi_n$ is defined as:
$$
\phi_{n}(x)=\left\{
\begin{aligned}
ax+b & , & l_1\le x < l_2 \\
\sigma(x) & , & otherwise 
\end{aligned}
\right.
$$

To quantify the quality of our approximation, we define the error metric for the entire FFN layer as:
$
err(x)=\sum_{n}^{h}err_{n}(x)
$
where each neuron's error contribution is measured using the L2 distance~\cite{Normmath13:online} between original and approximated outputs:
$
err_{n}(x)=||\sigma(xW_{1_{:,n}})W_{2_{n,:}}-\phi_n(xW_{1_{:,n}})W_{2_{n,:}}||_2
$

\noindent
\textbf{Adaptive Thresholding.}
Given the definition of FFN approximation error, \sys needs to determine appropriate linear approximation ranges for each neuron while satisfying the user-specified in-range threshold $t$. A key challenge is deciding how much of each neuron's input should be covered by the linear approximation.

Based on our insight in \Cref{sec:insight} that different layers and neurons contribute unequally to the model's performance, we develop an adaptive thresholding mechanism that assigns different coverage requirements to different components. This mechanism operates at two levels:

1) \textit{Layer-level thresholding}: For each layer $i$, we assign a threshold $t_i$ that determines what percentage of inputs should be linearly approximated. This is formulated as an optimization problem:
$$
\begin{aligned}
minimize\sum_{i}^{L}E_{i}t_{i},\ 
s.t.,\sum_{i}^{L}t_{{i}}=tL
\end{aligned}
$$
where $E_i$ represents the total empirical error measured on a calibration dataset for layer $i$, $L$ is the total number of layers, and $t$ is the target threshold. 
The empirical error $E_i$ is estimated by approximating all neurons in layer $i$ under threshold $t_i$ (e.g., as \Cref{fig:dist-errors-layer} shows).
The constraint ensures that the average threshold across layers equals the target.

2) \textit{Neuron-level thresholding}: Within each layer $i$, we further distribute the layer's threshold $t_i$ across its neurons using:
$$
\begin{aligned}
minimize\sum_{n}^{h}E_{i_n}t_{i_n},\ 
s.t.,\sum_{n}^{h}t_{i_n}=t_{i}h
\end{aligned}
$$
where $E_{i_n}$ represents the error contribution of neuron $n$ in layer $i$, $h$ is the number of neurons per layer, and $t_{i_n}$ is the threshold assigned to neuron $n$. This ensures more important neurons (those with higher error impact) receive more conservative thresholds.

This two-level adaptive thresholding approach allows \sys to maintain high approximation accuracy while satisfying user's threshold requirement. More critical layers and neurons are assigned stricter thresholds, while less important ones can use more aggressive linear approximations.

\begin{algorithm}[h]
\caption{Non-linear Function Approximation}
\label{alg:linear_range_generation}
\textbf{Input:} The non-linear function $\sigma$, threshold $t$, large language model $M$, calibration dataset $X$, range increment step $s$.\\
\textbf{Output:} Set of generated linear ranges $P$ for each neuron in each layer.\\
$err_{layers}\gets$estimate\_error\_layers($M.FFNs$, $X$)\\
\label{alg:linear_range_generation:layer_error}
$t_{layers}\gets$error\_aware\_threshold($err_{layers}$, $t$)\\
\label{alg:linear_range_generation:layer_threshold}
$P\gets \{\}$\\
\For{$t_i$ in $t_{layers}$}{
    $m\gets M.FFNs[i]$\\
    $err_{neurons}\gets$estimate\_error\_neurons($m$, $t_i$, $X$)\\
    $t_{neurons}\gets$error\_aware\_threshold($err_{neurons}$, $t_i$)\\
    $N\gets \{\}$\\
    \For{$t_{i_n}$ in $t_{neurons}$}{
        \label{alg:linear_range_generation:neural_threshold_start}
        \codecomment{/* use kernel density estimation to find a centroid */}\\
        $centroid\gets$find\_centroid($m.neuron_n$, $t_{i_n}$, $X$)\\
        \label{alg:linear_range_generation:find_centroid}
        \codecomment{/* initialize the range around centroid */}\\
        $R\gets$(left=centroid, right=centroid)\\
        \While{portion of input covered by $R$ < $t_{i_n}$}{
            $R_l\gets$($R$.left - s, $R$.left)\\
            \label{alg:linear_range_generation:step_start}
            $R_r\gets$($R$.left, $R$.right + s)\\
            \label{alg:linear_range_generation:step_end}
        \codecomment{/* approximate the range with a linear regressior, and evaluate the error */}\\
            $err_{left}\gets$approx\_error($m.neuron_n$, $R_l$, $X$)\\
            \label{alg:linear_range_generation:approx_error_start}
            $err_{right}\gets$approx\_error($m.neuron_n$, $R_r$, $X$)\\
            \label{alg:linear_range_generation:approx_error_end}
            \If{$err_{left} < err_{right}$}{
                \label{alg:linear_range_generation:choose_start}
                $R\gets R_l$\\
            }
            \Else{
                $R\gets R_r$\\
            }
            \label{alg:linear_range_generation:choose_end}   
        }
        $N[n]\gets R$\\
    }
    \label{alg:linear_range_generation:neural_threshold_end}
    $P[i]=N$\\
}
\Return{P}
\end{algorithm}

\noindent
\textbf{Range Searching.}
For each neuron, \sys employs a greedy search strategy to determine the optimal linear approximation range, following these steps. 
1) Initialize a starting range centered at the input distribution's centroid, computed using kernel density estimation (KDE)~\cite{terrell1992variable,parzen1962estimation}. 
2) Gradually expand the range boundaries until reaching the neuron-specific coverage threshold $t_{i_n}$.
3) During expansion, select the direction (left or right) that produces minimal approximation error.
4) Once the optimal range is determined, fit a linear function using least squares regression.
The complete algorithm is formalized in \Cref{alg:linear_range_generation}, leveraging a small calibration data to ensure both approximation accuracy and computational efficiency without fine-tuning the model.
\subsection{Constant Folded Matrix Generation}
\label{sec:matrix_gen}
Given the approximated linear ranges for each neuron in each layer, \sys generates constant folded matrices \(C\) and bias vectors \(B\) through a systematic transformation process. 

For each FFN layer, \sys first substitutes the activation function with the approximated linear function $\phi_n$ for each neuron $n$. This transformation creates a sequence of three consecutive linear operations per neuron: an initial linear transform (on $W_{1_{:,n}}$), followed by the approximated linear transform (on $\phi_n$ in $[l_1,l_2)$), and finally a second linear transform (on $W_{2_{n,:}}$). These three transformations are then constant folded into a single linear transformation.
The constant folding process for each neuron can be formally expressed as:
$$
\begin{aligned}
CF(W_{1_{:,n}}, W_{2_{n,:}}, \phi_{n})=(aW_{1_{:,n}}W_{2_{n,:}},bW_{2_{n,:}})=(C_n, B_n)
\end{aligned}
$$
Here, $CF$ represents the constant folding function for neuron \(n\), producing a matrix \(C_{n}\in\mathbb{R}^{d\times d}\) and a bias vector \(B_{n}\in\mathbb{R}^{1\times d}\). The parameters \(a\) and \(b\) denote the slope and intercept of the approximated linear function \(\phi_{n}\), respectively.
To complete the transformation for each layer, \sys summarizes all $h$ individual neuron's folded matrix into a single consolidated matrix \(C\in\mathbb{R}^{d\times d}\) and bias vector \(B\in\mathbb{R}^{1\times d}\):
$
(C, B)=(\sum_{n}^{h}C_n, \sum_{n}^{h}B_n)
$ for each layer, respectively.

\subsection{Predictor Generation}
\label{sec:predictor_gen}
While \Cref{sec:matrix_gen} demonstrates how to generate optimized matrices for linear approximation, we need a practical mechanism to handle cases where inputs fall outside the approximated range (\Cref{sec:insight}). 

To address this, \sys introduces a \textit{speculative approximation with result fixing} scheme. The key idea is to avoid expensive range checks before computation. Instead, \sys optimistically assumes all inputs fall within the linear range and proceeds with fast matrix operations. It then employs a lightweight predictor to identify which results need correction, only fixing the inaccurate ones (\Cref{sec:runtime}).

The predictor's role is straightforward - it takes the FFN block input and determines which neurons received inputs outside their approximated linear ranges. To implement this efficiently, \sys creates a compressed version of the weight matrix $W_1$, which contains just enough information to make these predictions without the overhead of full matrix operations.
For compressing $W_1$, \sys uses quantization techniques~\cite{frantar2023gptqaccurateposttrainingquantization}. While alternatives like training smaller prediction networks are possible~\cite{song2024powerinfer,liu2023deja}, quantization provides an effective balance of accuracy and simplicity. \sys can incorporate various LLM compression methods like pruning or knowledge distillation, as long as they preserve sufficient information to predict out-of-range inputs. This enables high performance while maintaining accuracy through selective result fixing.

\subsection{Inference Runtime}
\label{sec:runtime}
The \sys operates in two phases: an offline phase that generates the constant folded matrix (\Cref{sec:matrix_gen}) and predictor (\Cref{sec:predictor_gen}), and an online phase that performs inference through speculative approximation and result fixing.

\noindent
\textbf{Speculative Approximation.} During this step, \sys approximates the FFN computation using a pre-computed constant folded matrix. This simplifies the complex FFN computation into a simple matrix multiplication and addition operation: $FFN(x)\approx xC+B$. This approximation significantly reduces the computational complexity compared to the original FFN computation.

\noindent
\textbf{Result Fixing.} Not all neurons can be accurately approximated using the simplified computation above. The predictor identifies these cases, and \sys corrects the results for these specific neurons using the original computation method. This correction involves two steps. First, we remove the approximate results for the identified neurons. Then, we compute and add back their actual results using the original FFN computation.

\begin{figure}
    \centering
    \includegraphics[width=\columnwidth]{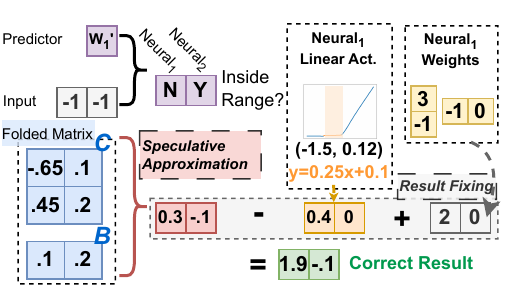}
    \caption{The speculative approximation and result fixing steps in the online phase of \sys.}
    \label{fig:speculate-and-fix}
\end{figure}

\Cref{fig:speculate-and-fix} illustrates this process with a concrete example. 
The system first generates an approximate result, then identifies and corrects inaccurate approximations by replacing them with precise computations.

\noindent
\textbf{Memory Footprint.} 
The memory requirements of \sys consist of three main components. 1) The constant folded matrix and its bias terms; 2) The lightweight predictor;
3) Only the original weights of neurons that require exact computation (fixing).
The system's memory efficiency varies between the two inference phases. In the first phase, more neurons might require their original weights, leading to higher memory usage. In contrast, during sequential token generation (second phase), fewer neurons typically need correction, resulting in better memory efficiency.

\section{Implementation}
\label{sec:impl}
We implemented \sys as a prototype using PyTorch~\cite{ReleaseP12:online}. 
We integrated \sys into the HuggingFace Transformer library (v4.41.0)~\cite{Transf37:online} and vLLM inference engine (v0.6.6)~\cite{vllmproj38:online}.
The system consists of two main components:

\noindent
\textbf{Offline Component.} The offline preprocessing pipeline, implemented in approximately 900 lines of Python code, handles non-linear function approximation and constant folding matrix generation. We leveraged cuML~\cite{Welcomet8:online}, a GPU-accelerated machine learning library, for training the linear regression model and performing kernel density estimation efficiently. The resulting constant-folded matrix is serialized to a binary format for runtime loading. The predictor is obtained by GPTQ~\cite{frantar2023gptqaccurateposttrainingquantization} with 2-bit precision, which is also stored in binary format, and loaded before inference.

\noindent
\textbf{Runtime Component.} The inference runtime consists of two parts: (1) a Python library implementing speculative approximation and result fixing logic (approximately 150 lines), and (2) a specialized NVIDIA CUDA kernel~\cite{CUDACPro24:online} for efficient result fixing. The kernel performs selective loading of FFN weights based on predicted neuron indices. We optimized the memory efficiency by memory coalescing~\cite{HowtoAcc72:online} and vectorized shared memory access~\cite{LDS128lo28:online}.

\section{Evaluation}
\label{sec:eval}
Our evaluation mainly focuses on the following questions:
\begin{itemize}
    \item How does \sys affect the model's performance in terms of perplexity and accuracy on downstream tasks?
    \item How does \sys compare with existing LLM pruning methods?
    \item How much pratical speedup can be gained during inference by \sys?
\end{itemize}
Apart from the above main questions, we evaluate the precision and effectiveness of the range assignment algorithm, predictor size's influence on model performance, inference runtime overhead, and the precision effects of reordering FFN.

\subsection{Settings}

\textbf{Models.} We evaluated \sys on five  large language models ranging from 1.6B to 11.1B parameters: Falcon-7B~\cite{almazrouei2023falconseriesopenlanguage}, Falcon2-11B~\cite{tiiuaefa57:online}, BLOOMZ-7B~\cite{le2023bloom}, GPT2-XL~\cite{openaico17:online}, and OPT-6.7B~\cite{facebook45:online}. These models represent different architectures and training objectives while being widely adopted in real-world applications. Each model employs a standard FFN architecture with either GELU or ReLU activation functions. 
\Cref{tab:model-statistics} summarizes the key characteristics of these models, including parameter count, architecture details, and activation functions.
\begin{table}[t]
  \Huge
  \centering
  \resizebox{\columnwidth}{!}{%
  \begin{tabular}{@{}lllll@{}}
  \toprule
  Name &
    Params. &
    \begin{tabular}[c]{@{}l@{}}\%Params. \\ in FFN\end{tabular} &
    \begin{tabular}[c]{@{}l@{}}Activation\\ Function\end{tabular} &
    \begin{tabular}[c]{@{}l@{}}\#Downloads\\ per month\end{tabular} \\ \midrule
  Falcon-7B  & 7.2B & 80\% & GELU & 75K  \\
  Falcon2-11B & 11.1B & 80\% & GELU & 19K  \\
  BLOOMZ-7B1 & 7.1B & 67\% & GELU & 9.3K \\
  GPT-2-XL   & 1.6B & 67\% & GELU & 306K \\
  OPT-6.7B   & 6.7B & 67\% & ReLU & 39K  \\ \bottomrule
  \end{tabular}%
  }
  \caption{Basic statistics of the evaluated LLMs. The third column shows the estimated percentage of total model parameters in the FFN layer.}
  \label{tab:model-statistics}
\end{table}


\noindent
\textbf{Evaluating Benchmarks.} Following prior works evaluating LLM performance~\cite{sun2023simple,lin2024awq,frantar2022optq,ma2023llm}, we evaluate models on language generation and zero-shot tasks with  lm-evaluation-harness~\cite{eval-harness}. 
For language generation, we measure perplexity on Wikitext-2~\cite{merity2016pointer}, C4~\cite{JMLR:v21:20-074}, and Penn Treebank (PTB)~\cite{10.3115/1075812.1075835}. 
On language generation tasks, perplexity measures how well a model predicts text by calculating the exponential of the average negative log-likelihood. Lower perplexity indicates better performance, with a perplexity of 1 representing perfect prediction.
For zero-shot evaluation, we test accuracy on PIQA~\cite{bisk2020piqa}, Lambada~\cite{Eleuther75:online}, and ARC-Challenge~\cite{allenaia10:online}, which cover reasoning, word prediction, and question answering tasks respectively. Higher accuracy indicates better model performance.

\noindent
\textbf{Setup.}
We conducted all our experiments on a single NVIDIA RTX 4090D GPU with 24GB memory.
Unless specified, the calibration dataset we used consists of 8 samples randomly choosen from the first shard of C4, each with 2048 tokens. C4 is a dataset consisting of web pages scraped from the internet, representing generic text data and making sure that we do not overfit to specifics tasks.

\noindent
\textbf{Baselines.} 
We compare \sys with Wanda~\cite{sun2023simple} and RIA~\cite{zhang2024plug}, two state-of-the-art pruning methods that prune models based on activation inputs and weights. In all experiments, we compress the FFN blocks with \sys or the baselines, while keeping the attention blocks intact. \sys's compression ratio is controlled by the threshold of input activations in the linear range (\Cref{sec:non_linear_approx}).  
We also account for the folded matrix and predictor size--around 10\% and 4.8\% of the model size--when calculating the compression ratio (\Cref{sec:runtime}). For baselines, the compression ratio is controlled by the pruning ratio, with pruned weights considered compressed.

\subsection{Folded Model's Performance}
We first quantifies the performance of the models folded by \sys on language generation tasks and zero-shot tasks, and compare with the baselines with varying compression ratios.

\noindent
\textbf{Summary.} 
Our evaluation reveals that \sys outperforms Wanda and RIA across different compression ratios:

1) At lower compression ratios (50\% and 70\%), \sys demonstrates marginally better performance in both language generation and zero-shot tasks, achieving 4.6× and 3.1× lower perplexity, 1.47× and 1.26× higher accuracy on average.

2) At higher compression ratio (80\%), \sys shows substantial improvements over the baselines, achieving up to 215× and 406× lower perplexity on average, while maintaining an average of 20.1\% and 19.1\% absolute improvement, with up to 65.7\% higher accuracy on zero-shot tasks.

\begin{figure}[ht]
    \centering
    \begin{subfigure}{\columnwidth}
        \centering
        \includegraphics[width=\columnwidth]{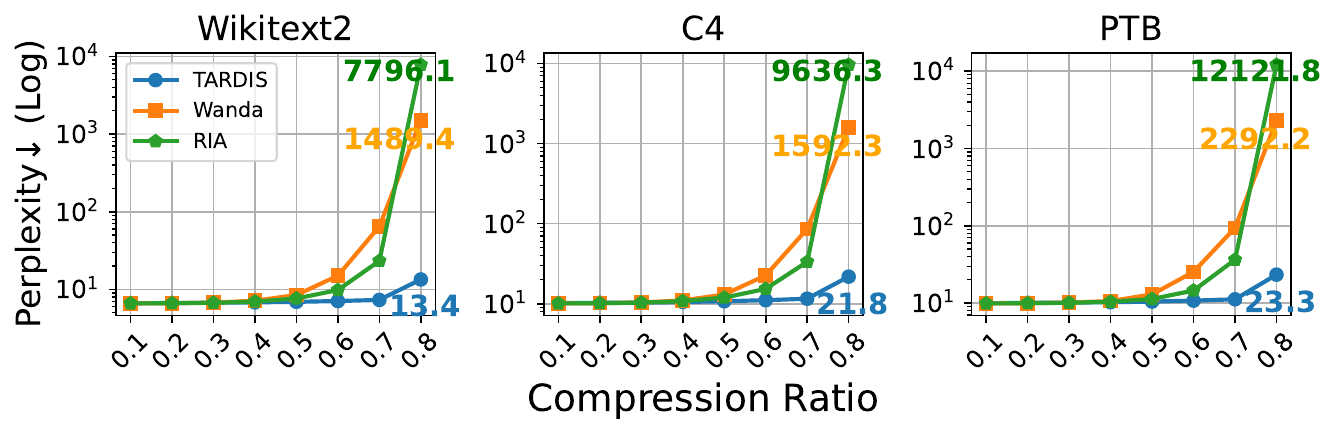}
        \caption{Perplexity (↓) comparison on WikiText-2, C4 and PDB.}
        \label{fig:perplexity-comparison}
    \end{subfigure}
    \begin{subfigure}{\columnwidth}
        \centering
        \includegraphics[width=\columnwidth]{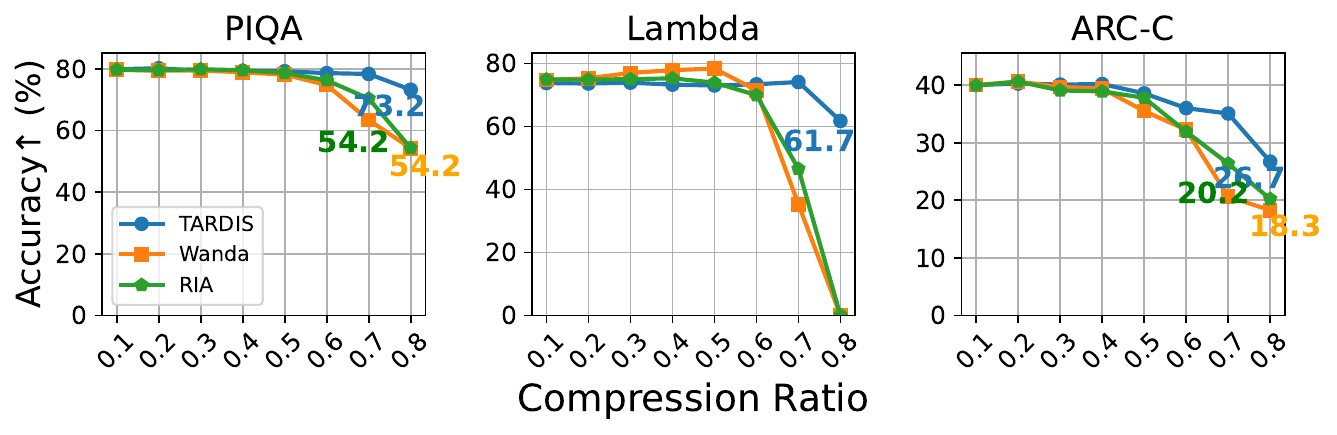}
        \caption{Accuracy (↑) comparison on PIQA, Lambada, and ARC-Challenge (ARC-C).}
        \label{fig:zero-shot-comparison}
    \end{subfigure}
    \caption{Performance comparison of Falcon-7B under different pruning methods and FFN compression ratios.}
    \label{fig:performance-comparison}
\end{figure}

\begin{table}[h]
  \Huge
  \centering
  \resizebox{\columnwidth}{!}{%
  \begin{tabular}{@{}lllllllllllccccc@{}}
  \toprule
  Dataset &
    Method &
    \multicolumn{3}{c}{Falcon-7B} &
    \multicolumn{3}{c}{Bloomz-7B1} &
    \multicolumn{3}{c}{GPT2-XL} &
    \multicolumn{3}{c}{OPT-6.7B} &
    \multicolumn{2}{c}{Falcon2-11B}\\ \midrule
  \multicolumn{2}{l}{\textit{Compress Ratio}} &
    \multicolumn{1}{c}{\textit{50\%}} &
    \multicolumn{1}{c}{\textit{70\%}} &
    \multicolumn{1}{c}{\textit{80\%}} &
    \multicolumn{1}{c}{\textit{50\%}} &
    \multicolumn{1}{c}{\textit{70\%}} &
    \multicolumn{1}{c}{\textit{80\%}} &
    \multicolumn{1}{c}{\textit{50\%}} &
    \multicolumn{1}{c}{\textit{70\%}} &
    \multicolumn{1}{c}{\textit{80\%}} &
    \textit{50\%} &
    \textit{70\%} &
    \textit{80\%} &
    \textit{70\%} &
    \textit{80\%} \\ \midrule
  Wikitext2 &
    Dense &
    \multicolumn{3}{c}{6.6} &
    \multicolumn{3}{c}{14.1} &
    \multicolumn{3}{c}{17.4} &
    \multicolumn{3}{c}{10.9}  &
    \multicolumn{2}{c}{5.2}\\
   &
    Wanda &
    8.4 &
    65.0 &
    1489 &
    16.6 &
    41.2 &
    467 &
    19.8 &
    175.1 &
    1.1e5 &
    11.2 &
    11.2 &
    8429 &
    57.3 &
    1715\\
   &
    RIA &
    7.3 &
    23.2 &
    7796 &
    15.9 &
    34.7 &
    481 &
    19.4 &
    107.8 &
    1.7e4 &
    \textbf{10.7} &
    72.0 &
    2.0e4 &
    25.4 &
    1427 \\
   &
    Ours &
    \textbf{6.9} &
    \textbf{7.3} &
    \textbf{13.4} &
    \textbf{14.9} &
    \textbf{16.2} &
    \textbf{387} &
    \textbf{19.3} &
    \textbf{39.9} &
    \textbf{313} &
    11.1 &
    \textbf{11.1} &
    \textbf{11.1} &
    \textbf{5.9} &
    \textbf{11.4} \\ \midrule
  C4 &
    Dense &
    \multicolumn{3}{c}{10.1} &
    \multicolumn{3}{c}{18.5} &
    \multicolumn{3}{c}{20.9} &
    \multicolumn{3}{c}{12.7} &
    \multicolumn{2}{c}{9.1} \\
   &
   Wanda &
   13.0 &
   85.8 &
   1592 &
   21.4 &
   51.4 &
   459 &
   23.7 &
   193.9 &
   9.0e4 &
   13.9 &
   62.9 &
   8274.5 &
   99.3 &
   2448 \\
  &
   RIA &
   11.9 &
   33.3 &
   9636 &
   21.0 &
   46.9 &
   812 &
   \textbf{23.4} &
   108.8 &
   1.5e4 &
   13.7 &
   106.8 &
   1.8e4 &
   38.7 &
   2016 \\
  &
   Ours &
   \textbf{10.7} &
   \textbf{11.7} &
   \textbf{21.8} &
   \textbf{19.1} &
   \textbf{20.3} &
   \textbf{456} &
   24.6 &
   \textbf{27.1} &
   \textbf{484} &
   \textbf{12.9} &
   \textbf{12.9} &
   \textbf{12.9} &
   \textbf{10.9} &
   \textbf{25} \\ \midrule
  PDB &
    Dense &
    \multicolumn{3}{c}{9.9} &
    \multicolumn{3}{c}{26.8} &
    \multicolumn{3}{c}{23.2} &
    \multicolumn{3}{c}{13.1} &
    \multicolumn{2}{c}{8.4} \\
 &
  Wanda &
  12.9 &
  93.9 &
  2292 &
  30.1 &
  87.1 &
  1058 &
  27.3 &
  185.0 &
  4.6e4 &
  17.2 &
  90.3 &
  6212 &
  106 &
  2631 \\
 &
  RIA &
  11.2 &
  36.4 &
  1.2e4 &
  30.3 &
  77.6 &
  3141 &
  \textbf{26.5} &
  129.5 &
  1.4e4 &
  14.6 &
  171.7 &
  1.2e4 &
  39.2 &
  2159 \\
 &
  Ours &
  \textbf{10.4} &
  \textbf{11.2} &
  \textbf{23.3} &
  \textbf{26.6} &
  \textbf{28.1} &
  \textbf{409} &
  30.8 &
  \textbf{34.7} &
  \textbf{268} &
  \textbf{13.4} &
  \textbf{13.4} &
  \textbf{13.4} &
  \textbf{9.5} &
  \textbf{20.1} \\ \bottomrule
  \end{tabular}%
  }
  \caption{Perplexity of different models on three dataset, when the model's FFN block is compressed with different methods. Text in bold marks the best.}
  \label{tab:ppl-all}
  \end{table}
\noindent
\textbf{Language Generation Tasks.}
\Cref{tab:ppl-all} shows \sys generally outperforms baselines. At 50\% compression ratio, \sys achieves comparable or slightly better perplexity (6.9 vs 7.3-8.4 for Falcon-7B on Wikitext2). At 70\% compression ratio, \sys's advantage becomes clear with significantly lower perplexity (7.3 vs 23.2-65.0 for Falcon-7B). At 80\%, \sys shows dramatic improvements - reducing perplexity from 1489 (Wanda) and 7796 (RIA) to 13.4 for Falcon-7B on Wikitext2. However, \sys occasionally performs worse, like with GPT2-XL on C4 at 50\% (24.6 vs 23.4), likely due to the small model size that enlarges approximation error.
For the OPT-6.7B model, results remain constant across all compression ratios since \sys assigns the same linear function in each case. This consistency stems from the model's activation pattern, where approximately 95\% of input values are negative and effectively reduced to zero, simplifying the approximation to a single linear function.
\Cref{fig:perplexity-comparison} demonstrates Falcon-7B's perplexity trends across fine-grained compression ratios (10-80\%), showing a clear advantage over baseline methods starting from 50\% compression ratio.

\begin{table}[ht]
  \Huge
  \centering
  \resizebox{\columnwidth}{!}{%
  \begin{tabular}{@{}lllllccccccccccc@{}}
  \toprule
  Dataset &
    Metohd &
    \multicolumn{3}{c}{Falcon-7B} &
    \multicolumn{3}{c}{Bloomz-7B1} &
    \multicolumn{3}{c}{GPT2-XL} &
    \multicolumn{3}{c}{OPT-6.7B} &
    \multicolumn{2}{c}{Falcon2-11B} \\ \midrule
  \multicolumn{2}{l}{\textit{Compress Ratio}} &
    \multicolumn{1}{c}{\textit{50\%}} &
    \multicolumn{1}{c}{\textit{70\%}} &
    \multicolumn{1}{c}{\textit{80\%}} &
    \textit{50\%} &
    \textit{70\%} &
    \textit{80\%} &
    \textit{50\%} &
    \textit{70\%} &
    \textit{80\%} &
    \textit{50\%} &
    \textit{70\%} &
    \textit{80\%} &
    \textit{70\%} &
    \textit{80\%} \\ \midrule
  PIQA &
    Dense &
    \multicolumn{3}{c}{79.5\%} &
    \multicolumn{3}{c}{76.4\%} &
    \multicolumn{3}{c}{70.8\%} &
    \multicolumn{3}{c}{76.3\%} &
    \multicolumn{2}{c}{80\%} \\
   &
    Wanda &
    78.1 &
    63.3 &
    54.1 &
    74.5 &
    67.7 &
    57.6 &
    69.4 &
    58.8 &
    54.6 &
    74.4 &
    58.2 &
    53.7 &
    64.5 &
    56.2 \\
   &
    RIA &
    78.6 &
    70.3 &
    54.2 &
    \textbf{74.6} &
    68.2 &
    60 &
    69.5 &
    62.0 &
    55.1 &
    75 &
    56.1 &
    53.9 &
    67.3 &
    55.4 \\
   &
    Ours &
    \textbf{79.1} &
    \textbf{78.4} &
    \textbf{73.1} &
    74.5 &
    \textbf{73.2} &
    \textbf{64.3} &
    \textbf{70.1} &
    \textbf{64.2} &
    \textbf{59} &
    \textbf{76.3} &
    \textbf{76.3} &
    \textbf{76.3} &
    \textbf{79.7} &
    \textbf{75.4} \\ \midrule
  Lambada &
    Dense &
    \multicolumn{3}{c}{74.5\%} &
    \multicolumn{3}{c}{55.9\%} &
    \multicolumn{3}{c}{51.1\%} &
    \multicolumn{3}{c}{67.6\%} &
    \multicolumn{2}{c}{73.5\%} \\
   &
    Wanda &
    \textbf{78.4} &
    35.3 &
    0 &
    57.8 &
    49.8 &
    13.4 &
    \textbf{53.8} &
    26.1 &
    0 &
    \textbf{70.1} &
    20 &
    0 &
    13.1 &
    0.2 \\
   &
    RIA &
    73.8 &
    46.5 &
    0 &
    56.6 &
    52.1 &
    \textbf{26.9} &
    51.1 &
    \textbf{30.4} &
    0 &
    \textbf{70.1} &
    12.1 &
    0 &
    27 &
    0 \\
   &
    Ours &
    73.0 &
    \textbf{74.3} &
    \textbf{61.7} &
    \textbf{51.5} &
    \textbf{49.4} &
    12.7 &
    42.9 &
    20.6 &
    \textbf{0.8} &
    68.1 &
    \textbf{68.1} &
    \textbf{68.1} &
    \textbf{73.2} &
    \textbf{65.7} \\ \midrule
  ARC-C &
    Dense &
    \multicolumn{3}{c}{40.3\%} &
    \multicolumn{3}{c}{40.3\%} &
    \multicolumn{3}{c}{25.1\%} &
    \multicolumn{3}{c}{30.5\%} &
    \multicolumn{2}{c}{50.1\%} \\
   &
    Wanda &
    35.5 &
    20.6 &
    18.3 &
    36.9 &
    28.8 &
    20.9 &
    22.9 &
    19.4 &
    23 &
    30.1 &
    20 &
    20.1 &
    26.9 &
    19.9 \\
   &
    RIA &
    37.8 &
    26.4 &
    20.2 &
    36.9 &
    29.4 &
    21.3 &
    \textbf{23.1} &
    18.9 &
    \textbf{21.2} &
    30 &
    20.8 &
    19.9 &
    28.2 &
    20 \\
   &
    Ours &
    \textbf{38.9} &
    \textbf{35.7} &
    \textbf{26.7} &
    \textbf{38.1} &
    \textbf{35.2} &
    \textbf{23.7} &
    22.5 &
    \textbf{19.4} &
    20.6 &
    \textbf{31.7} &
    \textbf{31.7} &
    \textbf{31.7} &
    \textbf{46.2} &
    \textbf{34.1} \\ \bottomrule
  \end{tabular}%
  }
  \caption{Zero-shot accuracy (\%) of downstream tasks on different models with varying compression ration. Text in bold marks the best.}
  \label{tab:zero-shot-all}
  \end{table}
\noindent
\textbf{Zero-Shot Tasks.}
\Cref{tab:zero-shot-all} shows advantages for \sys in most cases. At 50\% compression, \sys matches or slightly exceeds baseline accuracy (79.1\% vs 78.1-78.6\% for Falcon-7B on PIQA). The gap widens at 70\% compression (78.4\% vs 63.3-70.3\%). At 80\%, while baselines degrade to ~54\%, \sys maintains 73.1\% accuracy on PIQA. One exception is Lambada, where Wanda performs better at 50\% compression (78.4\% vs 73.0\%), likely due to its pruning approach. For OPT-6.7B, \sys achieves nearly lossless compress at even 80\%. 
For Falcon2-11B, it achieves an average of only 9.5\% accuracy drop on the highest compression ratio (80\%), while accuracy of Wanda and RIA drops by more than 42\% on average.
\Cref{fig:zero-shot-comparison} shows performance across compression ratios on Falcon-7B.
While Wanda achieves better results on Lambada between 30-50\% compression, \sys demonstrates superior performance at higher ratios.



\subsection{Range Assignment Algorithm}
Recall that the range assignment algorithm is used to determine the linear range for each neuron in the model (\Cref{sec:non_linear_approx}), which is one of the key components of \sys.
In this section, we evaluate the preciness, effectiveness, and sensitivity in terms of calibration dataset of the range assignment algorithm.

\begin{figure}[ht]
    \includegraphics[width=0.9\columnwidth]{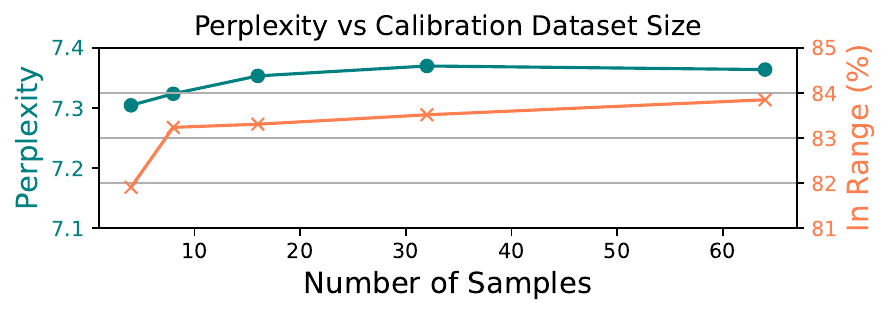}
    \centering
    \caption{Perplexity (left) and actual percentage of tokens with activation input in the linear range (right) versus the number of samples in the calibration dataset.}
    \label{fig:range-assignment-algorithm}
\end{figure}

\noindent
\textbf{Precision.}
We assess the precision of the range assignment algorithm by comparing the assigned range's actual threshold with the target threshold on the WikiText2 dataset. 
The right y-axis (in orange) of \Cref{fig:range-assignment-algorithm} profiles the actual percentage of tokens with activation inputs in the linear range, targeting 85\%. With as few as 8 samples, the actual percentage is close to the target (difference $<$1.8\%). As sample size increases, the percentage stabilizes and aligns more closely with the target.

\noindent
\textbf{Data-efficiency and Sensitivity on Calibration Set.}
\sys demonstrates high data efficiency, requiring only a small calibration dataset as it doesn't rely on backpropagation for weight updates. We evaluate this efficiency on Falcon-7B using WikiText-2. The left y-axis (in green) of \Cref{fig:range-assignment-algorithm} illustrates the model's perplexity across different calibration dataset sizes. The perplexity remains stable across 8 to 64 samples, varying by less than 0.06, indicating strong robustness to sample size. Notably, we observe that perplexity correlates strongly with the percentage of tokens having activation inputs in the linear range, as a higher percentage leads to increased cumulative approximation error.

To assess the calibration dataset's distribution sensitivity, we conducted cross-evaluation experiments using WikiText-2 and C4 subsets (8 samples each). Specifically, we calibrated \sys on one dataset and evaluated performance on the other for language generation tasks.
\begin{table}[ht]
    \centering
    \resizebox{0.55\columnwidth}{!}{%
    \begin{tabular}{@{}llll@{}}
    \toprule
    Eval\textbackslash{}Calib & Wikitext2 & C4    & Diff \\ \midrule
    Wikitext2                 & 7.26      & 7.34  & 0.08 \\
    C4                        & 11.7      & 11.33 & 0.37 \\ \bottomrule
    \end{tabular}%
    }
    \caption{Sensitivity of \sys to the calibration set distribution, in terms of perplexity.}
    \label{tab:sensitity-calib-dataset}
\end{table} 
As shown in \Cref{tab:sensitity-calib-dataset}, the choice of calibration dataset has minimal impact on performance, with perplexity variations of only 0.08 and 0.37 between different calibration sources.

\noindent
\textbf{Effectiveness.}
During offline analysis, our range assignment algorithm computes linear ranges for each neuron in the model. For Falcon-7B with 8 calibration samples and a target threshold of 0.85, the range search process takes approximately 30 minutes per layer, resulting in a total processing time of 16 hours for the complete model. The subsequent model folding step, which applies the identified linear ranges, is significantly faster, requiring only 6 seconds per layer. This one-time preprocessing cost is amortized across all future model inferences.

\subsection{Inference Speedup}
\begin{figure}
    \centering
    \includegraphics[width=\columnwidth]{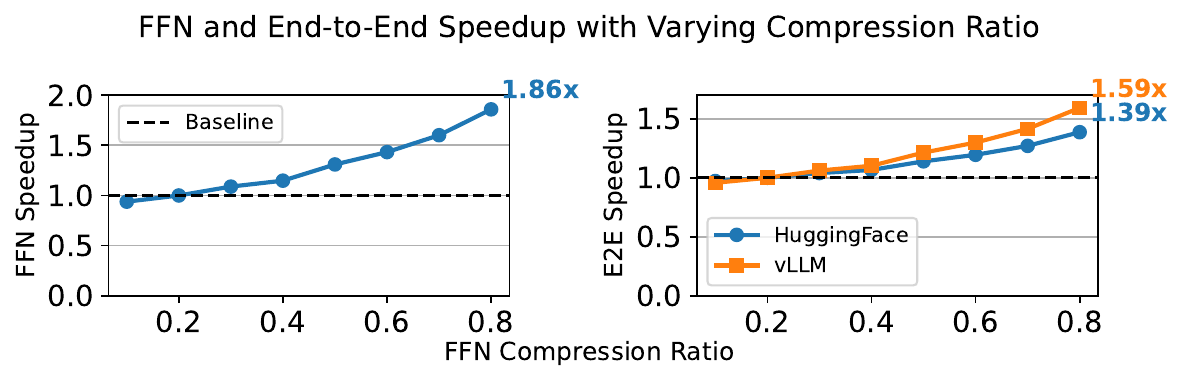}
    \caption{Inference Speedup of Falcon-7B with \sys}
    \label{fig:eval-speedup}
\end{figure}

We evaluate the practical inference speedup achieved by \sys on Falcon-7B across different compression ratios. We measure both FFN block-level and end-to-end speedup using the vLLM and HuggingFace frameworks. For evaluation, we perform text generation starting with 8 tokens and generating 192 additional tokens, comparing inference times across compression ratios from 10\% to 80\%. As shown in \Cref{fig:eval-speedup}, \sys begins providing inference benefits at 30\% compression. The speedup increases with higher compression ratios, reaching $1.86\times$ FFN speedup at 80\% compression, with end-to-end inference speedups of $1.39\times$ on HuggingFace and $1.59\times$ on vLLM. vLLM achieves better speedup due to its optimized attention mechanism~\cite{10.1145/3600006.3613165} - attention takes 55\% of inference time in HuggingFace but only 31\% in vLLM.

When processing sequences with many initial tokens but generating few output tokens (a less common scenario~\cite{philschm22:online}), \sys shows limited performance gains. 
This is because during the initial processing phase, each input token tends to activate different neurons, reducing the sparsity patterns that \sys leverages for acceleration.

\begin{figure}[ht]
    \centering
    \includegraphics[width=\columnwidth]{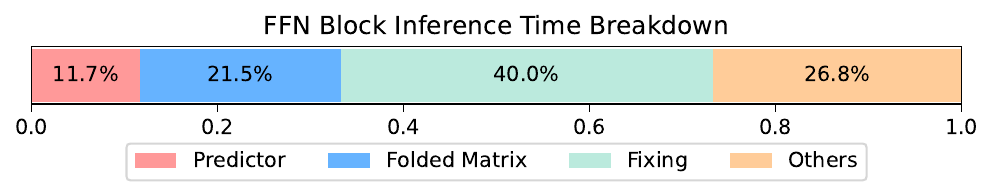}
    \caption{Performance breakdown of Falcon-7B FFN block during inference under a threshold of 0.85}
    \label{fig:ffn-block-breakdown}
\end{figure}
\subsection{Performance Breakdown}
We analyze the inference time breakdown of \sys by measuring four key components: predictor execution, constant folded matrix computation, result fixing, and auxiliary operations (e.g., mask generation and index conversion, which is used to select the wrongly assumed neurons). As shown in \Cref{fig:ffn-block-breakdown} with a target threshold of 0.85, result fixing dominates the execution time due to expensive sparse matrix indexing and loading operations in the CUDA kernel. The predictor accounts for approximately 12\% of the total time, while folded matrix computation comprises 22\%. The remaining time is spent on auxiliary operations, demonstrating that computational overhead primarily comes from sparse operations rather than the prediction mechanism.

\subsection{Influence of Predictor Size}
\begin{figure}[ht]
    \centering
    \includegraphics[width=\columnwidth]{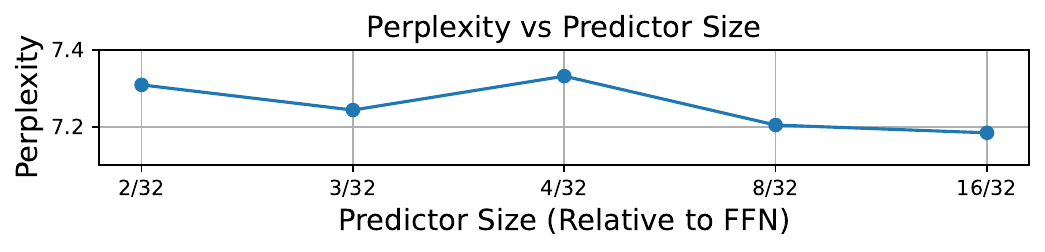}
    \caption{Influence of predictor size on perplexity in Wikitext2 dataset, evaluated with Falcon-7B.}
    \label{fig:predictor-size-vs-ppl}
\end{figure}

The predictor's accuracy in \sys, influenced by its size, affects model performance~\cite{song2024powerinfer}. We evaluated Falcon-7B's perplexity on WikiText2 with varying predictor sizes, controlled by the number of bits used in quantization~\cite{frantar2023gptqaccurateposttrainingquantization} of the weight matrix in FFN. \Cref{fig:predictor-size-vs-ppl} shows that perplexity decreases as predictor size increases, with a maximum difference of 0.12. This suggests that \sys can use a small predictor to balance performance and memory footprint.

\subsection{Precision Effects on Reordering FFN}
\label{sec:precision_affects}
Floating point arithmetic exhibits commutativity but not perfect associativity or distributivity due to rounding errors in finite precision operations~\cite{SixthCom51:online}. While \sys applies these properties during constant folding, we empirically evaluate their numerical impact. We conduct two experiments: first comparing the perplexity between folded and non-folded implementations, and second measuring the direct numerical differences in matrix operations on larger scales.

For the perplexity comparison, we evaluate three scenarios: (1) sequential computation without folding, (2) direct computation with the folded matrix, and (3) hybrid computation using ground truth activation inputs for linear range determination to isolate predictor effects.

\begin{table}[ht]
    \Large
    \centering
    \resizebox{\columnwidth}{!}{%
    \begin{tabular}{@{}llllll@{}}
    \toprule
    Intermediate Type & Original & bfloat16 & float16 & float32 & float64 \\ \midrule
    Average MSE       & 0        & 2.9e-3   & 4.9e-5  & 1.3e-5  & 1.3e-5  \\
    Perplexity        & 7.184    & 7.369    & 7.185   & 7.186   & 7.185   \\ \bottomrule
    \end{tabular}%
    }
    \caption{Perplexity and FFN MSE on WikiText2 with and without folding by using different intermediate data types during folded matrix generation}
    \label{tab:cf_intermediate_type_vs_ppl}
\end{table}
\Cref{tab:cf_intermediate_type_vs_ppl} presents the perplexity measurements across different floating-point precisions. The results show minimal differences between folded and non-folded implementations when using float32 or float16, with variations less than 0.1\%. Only bfloat16, known for its reduced precision in mantissa bits, shows notable differences.

\begin{table}[ht]
    \Large
    \centering
    \resizebox{0.55\columnwidth}{!}{%
    \begin{tabular}{@{}llll@{}}
    \toprule
    FFN Size & $\times$1   & $\times$4     & $\times$8     \\ \midrule
    MSE          & 1.7e-8 & 5.1e-7 & 1.5e-6 \\ \bottomrule
    \end{tabular}%
    }
    \caption{The MSE of different size of the constant folded matrix vs. original matrix in the first layer of the FFN in Falcon-7B. The intermediate datatype used in constant folding is float64.}
    \label{tab:mse_different_matrix}
\end{table}

To quantify the numerical stability across different scales, we computed the MSE between folded and original matrix computations. \Cref{tab:mse_different_matrix} presents MSE measurements for matrices at original Falcon-7B size ($1\times$) and manually scaled versions (enlarging the FFN by $4\times$ and $8\times$ larger). The consistently low MSE values ($<10^{-6}$) across all scales demonstrate that the numerical errors from matrix reordering remain negligible even as matrix dimensions increase.




\section{Related Works}
\label{sec:related}

\noindent
\textbf{LLM Compression.}
Recent LLM compression approaches fall into four categories:
1) Pruning removes weights through structural pruning~\cite{ma2023llm,zhang2024plug}, unstructural pruning~\cite{sun2023simple,liu2023deja}, and contextual pruning~\cite{song2024powerinfer,liu2023deja}. While effective, pruning often suffers from accuracy degradation at high compression ratios, and contextual pruning is mainly limited to ReLU-based models.
2) Quantization reduces parameter precision~\cite{frantar2022optq,lin2024awq,wei2023outlier,kim2023finequant} by converting floating-point weights to lower bit representations. 
3) Knowledge distillation transfers knowledge from large teacher models to smaller student models~\cite{gu2023knowledge,timiryasov2023baby}. 
4) Low-rank factorization decomposes weight matrices into products of smaller matrices~\cite{li2023losparse,saha2023matrix,chand2023dsformer}. 
The latter two approaches maintain good accuracy but require expensive training.

Unlike these direct parameter manipulation approaches, \sys introduces a novel compression paradigm by modifying activation functions to enable parameter reduction through constant folding. This achieves better accuracy under high compression ratios while maintaining model interpretability, and can be combined with existing compression methods for enhanced results.


\noindent
\textbf{Activation Function Approximation.}
Activation function approximation approaches are either hardware-centric or software-centric. Hardware approaches~\cite{zaki2019novel,mishra2007implementation} focus on FPGA implementations and simpler neural networks. Software approaches~\cite{liao2023method} target training optimization through piecewise linear approximation.

In contrast, \sys combines activation function approximation with constant folding, enabling parameter reduction through matrix multiplication merging for both memory and speed benefits.

\section{Discussion and Limitations}
The primary limitation of \sys lies in its handling of LLMs that use GLU-variant Feed-Forward Networks~\cite{shazeer2020gluvariantsimprovetransformer} (Gated-FFN), which make up about 53\% of the most popular text-generation models on HuggingFace. 
Models like LLaMA3.1~\cite{dubey2024llama} and Qwen 2.5~\cite{yang2024qwen2} use a FFN variant with quadratic operations: $FFN(x)=\sigma(xW_1)\odot (xW_2)W_3$, where $\odot$ is elementwise multiplication. This architecture poses significant challenges for matrix folding since the folded matrices grow exponentially. Applying our approach to LLaMA-2-7B would increase parameters by 254$\times$, undermining our goal of reducing computational costs. The Multi-head attention block encounters similar limitations.
Future works could focus on developing specialized folding techniques that can efficiently handle these quadratic operations while maintaining reasonable memory requirements.

\section{Conclusion}
We present \sys, a novel approach to compress LLMs
by applying constant folding optimization to feed-forward
networks. Our method approximates non-linear activations with linear functions in common input ranges, enabling parameter reduction through matrix multiplication reordering. Experiments show \sys reduces FFN parameters by up to 80\% while maintaining accuracy, achieving 1.6$\times$ inference speedup on vLLM and 1.4$\times$ on HuggingFace.

\bibliographystyle{plain}
\bibliography{paper}

\end{document}